\documentclass[runningheads]{llncs}

\usepackage[T1]{fontenc}
\usepackage{graphicx}
\usepackage{xcolor}
\usepackage{float}
\usepackage{tikz, pgfplots}
\usepackage{amsmath}
\usetikzlibrary{positioning}
\usepackage{hyperref}
\usepackage{subcaption}

\begin{document}

\title{The Return of Structural Handwritten Mathematical Expression Recognition}
\titlerunning{The Return of Structural HMER}

\author{Jakob Seitz \and Tobias Lengfeld \and Radu Timofte}
\authorrunning{J. Seitz, T. Lengfeld, R. Timofte}
\institute{Computer Vision Lab, CAIDAS \& IFI, University of W\"urzburg, W\"urzburg, Germany \\
\email{JakobSeitz007@gmail.com}, \email{tobias.lengfeld@web.de}, \email{radu.timofte@uni-wuerzburg.de}}

\maketitle

\begin{abstract}
Handwritten Mathematical Expression Recognition (HMER) is foundational for educational technologies, enabling applications like digital note-taking and automated grading. While modern encoder-decoder architectures with large language models excel at LaTeX generation, they lack explicit symbol-to-trace alignment – a critical limitation for error analysis, interpretability, and spatially aware interactive applications requiring selective content updates.

This paper introduces a structural recognition approach with two innovations: (1) an automatic annotation system that uses a neural network to map LaTeX equations to raw traces, automatically generating annotations for symbol segmentation, classification, and spatial relations, and (2) a modular structural recognition system that independently optimizes segmentation, classification, and relation prediction. By leveraging a dataset enriched with structural annotations from our auto-labeling system, the proposed recognition system combines graph-based trace sorting, a hybrid convolutional-recurrent network, and transformer-based correction to achieve competitive performance on the CROHME-2023 benchmark. Crucially, our structural recognition system generates a complete graph structure that directly links handwritten traces to predicted symbols, enabling transparent error analysis and interpretable outputs.

Our results challenge the perception that structural methods are outdated, demonstrating their viability when supported by high-quality annotated data. To promote reproducibility and future research, we release the annotated \href{https://doi.org/10.5281/zenodo.14968570}{CROHME+} and \href{https://doi.org/10.5281/zenodo.15091575}{MathWriting+} datasets, providing data for advancing interpretable HMER systems.

\end{abstract}

\textbf{CROHME+} dataset: \href{https://doi.org/10.5281/zenodo.14968570}{https://doi.org/10.5281/zenodo.14968570}

\textbf{MathWriting+} dataset: \href{https://doi.org/10.5281/zenodo.15091575}{https://doi.org/10.5281/zenodo.15091575}

\newpage

\section{Introduction}  
\label{sec:introduction}  

Handwritten Mathematical Expression Recognition (HMER) is essential for applications like educational tools, automated grading, and digital note-taking. The growing use of tablets demands robust systems for converting handwritten input into digital form. Yet, the complexity of mathematical notation - spatial hierarchies (e.g., superscripts, fractions), ambiguous symbols (e.g., $O$ vs. $0$), and diverse handwriting - poses major challenges.

Early systems used rule-based structural parsing \cite{anderson1967syntax,chang1970method}, while modern approaches rely on encoder-decoder models \cite{wap,tap} that map traces to LaTeX. Recent work integrates large language models (LLMs) \cite{tap,pal-v2} and graph-based reasoning \cite{g2g,xie2024local,truong2024survey}, improving accuracy but often lacking explicit trace-to-symbol alignment. This limits interpretability, error analysis, and spatially aware tasks.

\begin{figure}
    \centering
    \begin{tikzpicture}
        \node (before) at (-1,0) {\includegraphics[width=.45\linewidth]{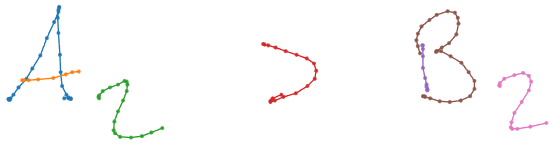}};
        \node[below, align=center] at (before.south) {(a) Raw trace data};
        \node (after) at (5.5,0) {\includegraphics[width=.45\linewidth]{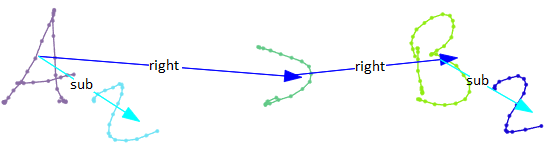}};
        \draw[->, thick] (before.east) -- (after.west);
        \node[below, align=center] at (after.south) {(b) Prediction: $A_2>B_2$};
    \end{tikzpicture}
    \caption{Task visualization: (a) Raw input traces of a handwritten expression; (b) Output of the proposed recognition system, showing a Stroke Label Graph (SLG) with segmented trace groups (colored traces) and spatial relationships (arrows), directly convertible to LaTeX.}
    \label{fig:pipeline}
\end{figure}
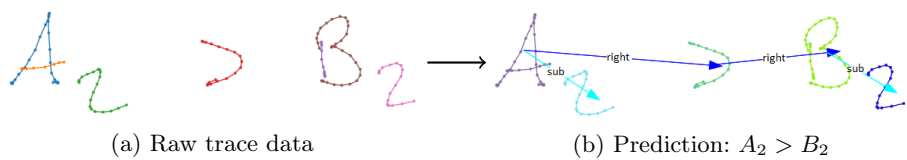

We address these limitations with two contributions:
\begin{itemize}
    \item \textbf{Automatic Annotation System:} We propose a neural model that aligns raw traces with LaTeX labels to automatically annotate 374,000 mathematical expressions. Our approach generates the essential MathML and TraceGroup annotations required to map traces to symbols within an equation, yielding two publicly available datasets -\href{https://doi.org/10.5281/zenodo.14968570}{CROHME+} and \href{https://doi.org/10.5281/zenodo.15091575}{MathWriting+} - that provide comprehensive, trace-level details to drive further research in online handwritten mathematical expression recognition.
    
    \item \textbf{Novel Structural Recognition System:} Our approach decouples HMER into segmentation, symbol classification, and relation classification for independent subtask optimization and transparent error analysis. Leveraging the additional annotations from CROHME+, our pipeline not only predicts LaTeX equations with competitive expression accuracy (74.14\%) but also produces a Stroke Label Graph (see Figure~\ref{fig:pipeline}), enabling use cases that depend on the spatial structure of the recognized equation.
\end{itemize}

This paper reviews related work (Section~\ref{sec:related_work}), details our annotation approach (Section~\ref{sec:structural_annotation}), presents the recognition system (Section~\ref{sec:recognition}), and discusses strengths and limitations (Section~\ref{sec:discussion}).

\section{Related Work}
\label{sec:related_work}

Handwritten Mathematical Expression Recognition (HMER) has advanced with deep learning, structured representations, and improved benchmarks. This section reviews key datasets and recognition methods.

\subsection{Benchmark Datasets}
\label{ssec:benchmark_datasets}

The \textbf{CROHME} dataset \cite{crohme2023} is the primary OHMER benchmark, providing diverse expressions and refined evaluation metrics. It uses the InkML format \cite{inkml} to organize trace data, but only 8.2\% of its training set (the self-collected portion) contains relational MathML annotations \cite{mathml}. Moreover, 44\% of samples lack reliable Trace Group labels, limiting their use for segmentation and classification.

Additional datasets cover specific aspects of HMER: \textbf{MathWriting} \cite{mathwriting} offers 254 symbol classes and many expressions but no relational or Trace Group labels; \textbf{Detexify} \cite{detexify} targets isolated symbol classification; and \textbf{IAMonDo} \cite{iamondo} integrates mathematical expressions within mixed-content documents. However, none provide the detailed trace-level grouping and hierarchical labels essential for modular, structural recognition systems.

\subsection{Recognition Approaches}
\label{ssec:recognition_approaches}

Early HMER systems relied on rule-based structural parsers \cite{anderson1967syntax,chang1970method,belaid1984syntactic}, which handled simple formulas but struggled with handwriting diversity and complex spatial hierarchies.

The advent of deep learning shifted the focus to \textbf{encoder-decoder architectures} that map raw traces directly to LaTeX. Models such as WAP \cite{wap}, BTTR \cite{bttr}, and CoMER \cite{comer} employ convolutional encoders with attention-based decoders. LLM-enhanced methods like TAP \cite{tap} and PAL-v2 \cite{pal-v2} further refine output through temporal modeling or syntactic constraints. These systems achieve high expression accuracy but lack explicit trace-to-symbol alignment and structural interpretability.

To address structural alignment, graph-based reasoning has been integrated into HMER. Wu et al. introduced the Graph-to-Graph (G2G) model for online HMER \cite{wu2021graph}, while Xie et al. proposed the GGM-EGAT network with local and global graph attention \cite{xie2024local}. Both enhance structural reasoning but jointly optimize segmentation, classification, and relation prediction, limiting targeted control.

Explicit structural recognition systems decompose these tasks to preserve trace-symbol alignment and produce interpretable outputs. Notable offline examples include QD-GGA \cite{mahdavi2020visual} and LGAP \cite{shah2023line}, which generate Symbol Layout Trees using graph-based parsing from images or connected components.

Our work extends structural recognition to the online domain using time-ordered stroke data with a modular, stage-by-stage pipeline.

\subsection{Summary}  
Deep learning has advanced OHMER from rule-based systems to encoder-decoder models with strong LaTeX generation, further improved by LLM post-correction. However, most approaches lack trace-to-symbol alignment and interpretable structural outputs. Graph-based methods add structural reasoning but often rely on joint optimization and limited annotations. Our work addresses these gaps through automated structural annotation and a modular, stage-by-stage online recognition system.

\section{Automating Structural Annotations}
\label{sec:structural_annotation}

Datasets like CROHME-2023 \cite{crohme2023} and MathWriting \cite{mathwriting} provide valuable data for OHMER, but lack comprehensive structural annotations. In CROHME, 92\% of the training set (synthetic data) lacks MathML, and many trace group labels are incorrect. MathWriting contains no MathML or trace group information, limiting its use for structure-dependent recognition systems.

\subsection{Proposed Annotation Network}

To address these limitations, we propose a neural network that enriches raw trace data with structural annotations. Given trace coordinates and a corresponding LaTeX label, the network predicts relation graphs and trace group information. This bridges the gap between high-level LaTeX and fine-grained structural annotations, simplifying dataset preparation for recognition tasks.

\subsubsection{Input Description}

The network input combines raw trace data and structural information extracted from the LaTeX label, effectively reducing the task to a segmentation problem.

\paragraph{Traces}  
Traces capture the pen trajectory of handwritten expressions as $(x, y)$ coordinates. Although some datasets include timestamps $(t)$, most CROHME samples lack temporal information, so only $(x, y)$ are used.

\paragraph{Structural Information}
Structural information from the LaTeX label defines symbol identities and relationships. For each step, the network receives the \textbf{reference symbol} $s_{\text{ref}}$, the \textbf{next symbol} $s_{\text{next}}$ with its relation $R(s_{\text{ref}}, s_{\text{next}})$, and the \textbf{neighbours} $\mathcal{N}(s_{\text{next}})$ with their respective relations $R(s_{\text{next}}, s_{\text{neigh}})$. This allows iterative mapping of LaTeX content to the underlying trace groups.

\subsubsection{Model Architecture}

The network architecture consists of a Bidirectional Long Short-Term Memory (BiLSTM), followed by fully connected layers. The BiLSTM processes the sequential trace data, capturing dependencies between symbols, while the fully connected layers refine the predictions. The final output is a binary segmentation mask, where each value indicates whether a trace belongs to the current symbol (1) or not (0). This architecture enables the network to map classification and relation information from a LaTeX expression to a binary segmentation mask, effectively predicting the grouping of traces into symbols.

\subsubsection{Training Process}

The network is trained on the CROHME-2023 training data, with the validation set used for early stopping and hyperparameter optimization. The test set is reserved for final evaluation. To improve generalization and reduce overfitting, the training process includes \textbf{dynamic data augmentation}, which applies transformations such as scaling, shearing, rotation, and translation during each epoch. Additionally, \textbf{dropout and layer regularization} are regularly applied to mitigate overfitting.

\subsection{Cross-Checking}

For validation, the auto-generated SLGs were filtered using ground-truth and model-based criteria to ensure high annotation quality. In CROHME, predicted segmentations had to exactly match available Trace Group labels; for samples without ground-truth grouping (e.g., synthetic data), each predicted symbol’s class had to appear in the classifier’s top-10 outputs. For MathWriting, which lacks ground-truth annotations, we applied a stricter re-check: each symbol was reclassified, and expressions were discarded if the new top prediction did not match the LaTeX label. These conservative filters ensured high-confidence annotations in CROHME+ and MathWriting+, though they may exclude rare patterns, such as uncommon symbols or unconventional notations, even when the SLG was correct.

\begin{figure}[H]
    \centering
    \begin{subfigure}[b]{0.45\textwidth}
        \centering
        \begin{tikzpicture}
                \begin{axis}[
                ybar stacked,
                width=1.1\textwidth, 
                height=5cm,
                symbolic x coords={Val, Test, Train, Synth},
                xtick=data,
                ymin=0,
                ymax=250000,
                ylabel={Number of Expressions},
                enlarge x limits=0.15,
                bar width=15pt,
                font=\small    
            ]
                \addplot+[ybar,fill={rgb,255:red,84;green,162;blue,75},draw opacity=0] plot coordinates {(Train,12000) (Synth,0) (Val,1704) (Test,3501)};
            
                \addplot+[ybar,fill={rgb,255:red,76;green,120;blue,168},draw opacity=0] plot coordinates {(Train,10) (Synth,131411) (Val,0) (Test,0)};
            
                \addplot+[ybar,fill={rgb,255:red,245;green,133;blue,24},draw opacity=0] plot coordinates {(Train,20) (Synth,13695) (Val,0) (Test,0)};
   
            \end{axis}
        \end{tikzpicture}
        \caption{CROHME}
        \label{fig:crohme_bar_chart}
    \end{subfigure}
    \hfill
    \begin{subfigure}[b]{0.5\textwidth}
        \centering
        \begin{tikzpicture}
                \begin{axis}[
                ybar stacked,
                width=1.1\textwidth, 
                height=5cm,
                symbolic x coords={Symbol, Test, Val, Synth, Train},
                xtick=data,
                ymin=0,
                ymax=250000,
                enlarge x limits=0.15,
                bar width=15pt,
                font=\small
            ]
                \addplot+[ybar,fill={rgb,255:red,84;green,162;blue,75},draw opacity=0] plot coordinates {(Symbol,0) (Test,0) (Val,0) (Synth,0) (Train,0)};
        
                \addplot+[ybar,fill={rgb,255:red,76;green,120;blue,168},draw opacity=0] plot coordinates {(Symbol,6300) (Test,6000) (Val,9000) (Synth,86000)(Train,143000) };
        
                \addplot+[ybar,fill={rgb,255:red,245;green,133;blue,24},draw opacity=0] plot coordinates {(Symbol,100) (Test,1600) (Val,7000) (Synth,35000) (Train,87000)};
            \end{axis}
        \end{tikzpicture}
        \caption{MathWriting}
        \label{fig:mathwriting_bar_chart}
    \end{subfigure}
    
    \vspace{-20pt}
    \begin{table}[H]
    \centering
    \begin{tabular}{|c c|}
    \hline
    \textcolor[rgb]{0.33, 0.63, 0.29}{\rule{2mm}{2mm}} & With Annotation \\
    \textcolor[rgb]{0.30, 0.47, 0.66}{\rule{2mm}{2mm}} & Generated Annotation \\
    \textcolor[rgb]{0.96, 0.52, 0.09}{\rule{2mm}{2mm}} & Without Annotation \\
    \hline
    \end{tabular}
    \end{table}
    \vspace{-25pt}
    
    \caption{Annotated and total expression counts for CROHME and MathWriting subsets (see \href{https://doi.org/10.5281/zenodo.14968570}{CROHME+} and \href{https://doi.org/10.5281/zenodo.15091575}{MathWriting+} dataset for more details).}
    \label{fig:bar_charts}
\end{figure}

\subsection{Results}

The graph labeling network achieved 99.45\% per-symbol accuracy on CROHME validation data, successfully annotating 84\% of the synthetically generated CROHME 2023 set and 66\% of MathWriting samples. Rigorous cross-checking ensured that only correctly labeled equations were retained, resulting in the enhanced \href{https://doi.org/10.5281/zenodo.14968570}{CROHME+} and \href{https://doi.org/10.5281/zenodo.15091575}{MathWriting+} datasets, which provide comprehensive Trace Group and relational annotations in MathML format. These resources significantly expand structurally annotated data for HMER while eliminating manual labeling efforts and enabling interpretable system training and streamlined error analysis.

Notably, the network generalized to the MathWriting dataset, which features 254 symbol classes compared to 101 in CROHME-2023 and lacks structural annotations entirely. Despite being trained exclusively on CROHME-2023, the network correctly labeled approximately 67\% of MathWriting's 374,177 equations. This cross-dataset generalization was enabled by leveraging similarities in symbol appearance and relational context within the surrounding expression structure.

Figure~\ref{fig:bar_charts} provides a detailed breakdown of the annotated data.

\section{Proposed Recognition System}
\label{sec:recognition}

Modern encoder-decoder systems often prioritize LaTeX generation over interpretability, limiting their ability to model spatial hierarchies. We propose a hybrid framework that combines deep learning with Stroke Label Graphs (SLGs), leveraging 133,000 SLG-labeled equations from the augmented CROHME dataset. The modular design decomposes recognition into five task-specific stages, optimized through the SLG structure.

\begin{figure}[h]
\centering
\includegraphics[width=\textwidth]{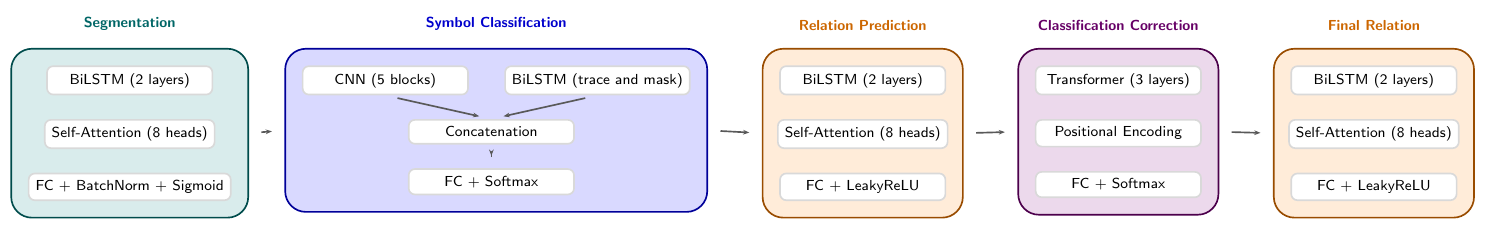}
\caption{System overview with five stages: (1) trace segmentation, (2) dual-modal classification, (3) primary relation prediction, (4) classification correction, (5) final relation prediction. Modules are trained independently but executed sequentially to produce a complete SLG convertible to LaTeX.}
\label{fig:architecture}
\end{figure}

The pipeline consists of:  
\begin{enumerate}
\item \textbf{Trace Segmentation}: Graph-based preprocessing and neural sequence modeling partition traces into symbol candidates (Section~\ref{sec:segmentation}).
\item \textbf{Dual-Modal Classification}: Trace trajectories (BiLSTM) and rendered glyphs (CNN) jointly predict symbol classes (Section~\ref{local_class}).
\item \textbf{Primary Relation Prediction}: An attention-recurrent network infers spatial dependencies from geometry and class labels (Section~\ref{sec:relation}).
\item \textbf{Global Context Correction}: A transformer refines symbol predictions using structural context from the SLG (Section~\ref{class_corr}).
\item \textbf{Revised Relation Prediction}: Final spatial relations complete the SLG via constrained tree generation (Section~\ref{sec:relation}).
\end{enumerate}

While modules are trained separately, they operate sequentially during inference to produce coherent SLGs directly convertible to LaTeX. The design supports targeted optimization and interpretable outputs, combining deep learning with graph-based reasoning for scalable, transparent expression recognition.

\subsection{Proposed Segmentation}
\label{sec:segmentation}

Segmentation forms the critical first step in our pipeline, decomposing raw trace data into discrete symbols that form the building blocks of mathematical expressions. Unlike offline image-based segmentation, our online approach leverages trace sequence data and spatial relationships to address challenges such as symbol overlaps, variable writing styles, and fragmented traces. The segmentation network achieves state-of-the-art accuracy through a combination of graph-based preprocessing and neural sequence modeling.

\subsubsection{Preprocessing}
The preprocessing pipeline transforms raw traces into a normalized representation optimized for segmentation through a series of structured steps. First, traces are processed in \textbf{reverse chronological order}, following a right-to-left spatial sequence that inversely mirrors the left-to-right temporal progression of handwritten equation composition. This approach operationalizes the observation that residual traces during \textbf{incremental segmentation} typically correspond to symbols authored earlier in the writing process. 

Next, a \textbf{topological graph} is constructed to model the relationships between traces. This involves creating a weighted undirected graph \(\mathcal{G}=(\mathcal{V},\mathcal{E})\), where vertices represent traces and edge weights are computed as minimal Euclidean distances using a \textbf{kdTree}. The \textbf{minimum spanning tree (MST)} is then extracted using Prim's algorithm to identify the 20 nearest neighbors to the rightmost trace. This method achieves a higher accuracy of \(98.03\%\) compared to \(96.89\%\) with direct Euclidean sorting, albeit at a higher computational cost, with a runtime of \(1.40\)ms versus \(0.04\)ms per symbol on a consumer PC. The complexity of this step is \(O(|\mathcal{E}| + |\mathcal{V}|\log|\mathcal{V}|)\).

Finally, \textbf{spatiotemporal normalization} is applied to standardize the trace data. This involves three key operations: \textbf{alignment}, \textbf{scaling}, and \textbf{resampling}. During alignment, the rightmost point of each trace is translated to the origin. Scaling then normalizes the coordinates to fit within the unit hypercube \([0,1]^2\). Resampling enforces uniform sampling density along the path using \textbf{piecewise linear interpolation}, defined by the equation 
\[
\mathbf{p}_i' = \mathbf{p}_k + \frac{s_i - s_k}{s_{k+1} - s_k} (\mathbf{p}_{k+1} - \mathbf{p}_k), \quad \text{where } s_k \leq s_i < s_{k+1}.
\]
The original \(M\) points are resampled equidistant along the total path length, ensuring a consistent representation for further processing.

\subsubsection{Network Architecture and Training}  
The segmentation network processes preprocessed traces through sequential and attention-based layers. Inputs consist of normalized \((x, y)\) coordinates from single traces. A two-layer Bidirectional LSTM (BiLSTM) models local trace dependencies, followed by an 8-head self-attention mechanism that captures global spatial relationships. The network outputs a binary segmentation mask via fully connected layers with sigmoid activation.  

Training employs a three-stage strategy. First, the \textbf{loss function} uses weighted binary cross-entropy (\(w=5\) for foreground) to counteract class imbalance. Second, \textbf{data augmentation} is applied to simulate handwriting variations through stochastic affine transformations. Scaling combines non-uniform (\(\sigma=0.2\)) and uniform (\(\sigma=0.4\)) factors, clamped to the range \([0.2, 5]\). Shear (\(\sigma_{x,y}=0.1\)), rotation (\(\sigma=8^\circ\)), and translation (\(\sigma_{x,y}=0.15\)) are sampled from normal distributions. These transformations preserve topological integrity while improving robustness. Third, \textbf{optimization} is performed using \textbf{AdamW} (\(\beta_1=0.9\), \(\beta_2=0.999\), \(\text{lr}=10^{-4}\)) with gradient clipping (\(||\nabla||_2 \leq 5\)) and early stopping to ensure stable convergence.

\subsubsection{Results and Analysis}

In summary, our segmentation approach, enhanced with synthetic labels and an MST-based sorting method, substantially improves performance in decomposing raw trace data into discrete symbols. To assess the impact of additional training data, we compared a model trained solely on CROHME-2023 real data with one having access to more data augmented using synthetic Trace Group annotations from our graph labeling network (see Section~\ref{sec:structural_annotation}).

The incorporation of more data elevated the symbol wise segmentation accuracy (Seg) from 98.22\% to 99.75\%, highlighting the benefits of enriched annotation. Notably, these enhancements also led to an increase in expression recognition accuracy (Exp) from 53.05\% to 56.64\% and structure recognition rate (Stru) from 75.02\% to 80.79\%. Our MST-based sorting technique achieves a Seg of 98.22\%, outperforming direct Euclidean sorting, which attains only 97.11\%. Although this method requires a significantly higher per-symbol runtime, the improved precision justifies the additional computational cost. Overall, these improvements make our approach 1.36\% more accurate than GGM-EGAT in terms of segmentation accuracy (see Table~\ref{tab:task_accuracy}).

\subsection{Proposed Symbol Classification}  
\label{sec:classification}  

Symbol classification, the highest-risk stage in OHMER, involves multiclass differentiation across 101 symbol categories (CROHME-2023), addressing challenges like glyph ambiguity (e.g., periods vs. commas), cross-cultural script variance, and high-dimensional class separation. We deploy a multimodal fusion framework: bidirectional LSTMs process sequential trace data to model temporal dependencies, while CNNs extract spatial features from rendered grayscale glyphs. A relation classification stage then predicts spatial dependencies between symbols, refining local predictions. Finally, a context-aware transformer encoder analyzes global structural relationships within the partially constructed SLG, resolving ambiguities through attention-based alignment of symbol semantics and layout topology. This hierarchical approach - combining local trace dynamics, spatial glyph morphology, and global syntactic constraints - ensures robust disambiguation, even for symbols with overlapping geometric signatures or context-dependent interpretations.

\subsubsection{Preprocessing}
\label{ssec:classification_preprocess}

The preprocessing pipeline optimizes input representations for multimodal classification through three key steps. First, \textbf{spatial normalization} is applied to standardize the input data. This involves centroid alignment to the origin, uniform scaling to a unit bounding box, and rotation correction based on the convex hull. These transformations ensure consistent spatial properties across inputs.

Second, \textbf{contextual rendering} is performed to capture the surrounding context of the traces. This includes rasterization into \(100 \times 100\) grayscale images, with neighboring traces rendered at reduced opacity (50\%). This step preserves the spatial relationships between traces while emphasizing the primary trace.

Third, \textbf{structural prior injection} is used to incorporate abstract information about already classified symbols. A 9-feature binary mask encodes symbol categories (e.g., whether a symbol is a variable), providing the model with additional structural context to improve classification accuracy.

\subsubsection{Local Classification Network}
\label{local_class}

The local classification framework employs a dual-stream neural network architecture, designed to process two distinct input modalities in parallel: (1) trace trajectories (temporal data) and (2) rendered images (spatial data). The architecture is optimized for robust feature extraction and fusion, as described below.

The \textbf{BiLSTM Pathway} processes interpolated trace coordinates in the temporal stream. It consists of three bidirectional LSTM layers (hidden size \(h = 256\)) to capture long-range temporal dependencies. Layer normalization is applied to stabilize hidden states, and a self-attention mechanism is incorporated to weight significant temporal features.

The \textbf{CNN Pathway} processes rendered images in the spatial stream using a 5-block VGG-style network. Each block includes convolutional layers with \(3 \times 3\) kernels and stride 1, batch normalization followed by ReLU activation, and max pooling with \(2 \times 2\) kernels and stride 2. The final feature map is flattened and projected to a fixed-size representation.

The \textbf{Fusion Layer} combines the outputs from the BiLSTM and CNN pathways by concatenating them into a unified feature vector. This vector is passed through three fully connected layers (512, 256, and 101 units) with ReLU activations and dropout (\(p = 0.4\)) to prevent overfitting.

\paragraph{Training Strategy:}
The dual-stream network is trained jointly in an end-to-end manner, optimizing both pathways simultaneously to ensure robust convergence and generalization. The training process employs a phased strategy with the following components.

The \textbf{loss function} is designed to mitigate class imbalance. Initially, the network uses a class-balanced cross-entropy loss, where each class weight \( w_c = 1 / f_c \) is inversely proportional to its frequency \( f_c \) in the training set. This ensures that underrepresented classes are adequately emphasized during the initial training phase. Once the model converges under this weighted scheme, we transition to unweighted standard cross-entropy (SCE) loss. This prevents the model from over-prioritizing rare classes and allows it to naturally favor more probable classes, aligning with the underlying data distribution.

For \textbf{optimization}, the network employs AdamW (\(\text{lr} = 10^{-6}\), weight decay \(\lambda = 10^{-4}\)) with gradient clipping (\(||\nabla||_2 \leq 5\)) and cosine annealing. Early stopping is applied based on validation loss to prevent overfitting and ensure stable convergence.

\subsubsection{Classification Correction}  
\label{class_corr}

The classification correction module refines the initial local symbol predictions by exploiting global structural dependencies across the expression. This transformer-based architecture processes a syntactically ordered sequence of classification probabilities and their trace information to resolve ambiguities and enforce syntactic consistency. The network comprises a 3-layer transformer encoder ($d_{\text{model}}=256$, 8 attention heads) that jointly processes symbol sequences and corresponding trace information. Each input is first projected via a learned linear embedding and augmented with additive absolute positional encodings:
\[
\mathbf{E}_t = \mathbf{W}_e \mathbf{x}_t + \mathbf{p}_t,
\]
where $\mathbf{x}_t$ denotes the input at position $t$, $\mathbf{W}_e$ is the embedding matrix, and $\mathbf{p}_t$ represents the trainable positional encoding. Each transformer layer employs pre-norm multi-head self-attention and position-wise feed-forward sublayers with residual connections. A subsequent two-layer classifier ($256 \rightarrow 128 \rightarrow C$) maps the final encoder states to class logits using leaky ReLU activations and dropout ($p=0.1$) for regularization.

Training is conducted in two phases. Initially, heavy data augmentation is applied to both spatial features and prediction probabilities, augmented further by synthetically generated data to enhance robustness and generalization. In a subsequent fine-tuning stage, the network is trained for a few epochs exclusively on real data without augmentation, thereby optimizing classification accuracy. The final output comprises contextually corrected classification probabilities for each symbol.

\subsubsection{Results and Analysis}

In summary, our multimodal fusion framework for symbol classification — which integrates temporal, spatial, and global contextual cues — substantially improves both symbol and expression recognition.
Symbol classification accuracy (Sym) assesses the percentage of correctly grouped traces with correct labels assigned at the symbol level.

Initially, an LSTM-based model trained solely on real data achieved a Sym of 92.53\% (Exp: 53.05\%). Augmenting the training set with synthetic labels raised Sym to 95.01\% (Exp: 64.08\%). Incorporating a CNN pathway further enhanced performance, increasing Sym to 96.55\% (a 1.54\% gain; Exp: 72.60\%). Finally, the application of a classification correction module that leverages a transformer encoder to incorporate global context refined Sym to 96.73\% (an additional 0.18\% gain; Exp: 74.27\%). These results underscore the effectiveness of our hierarchical approach in resolving symbol ambiguities and achieving robust recognition performance. Overall, these enhancements make our approach 5.96\% more accurate in symbol classification than GGM-EGAT (see Table~\ref{tab:task_accuracy}).

\subsection{Proposed Symbol Relation}  
\label{sec:relation}  

The symbol relation module constitutes the structural backbone of OHMER, tasked with inferring hierarchical dependencies between classified symbols to construct a parseable SLG. This stage faces three critical challenges: 1) disambiguating spatially overlapping symbols (e.g., distinguishing subscript vs. right relations), 2) resolving implicit grouping in absence of explicit brackets, and 3) maintaining consistency across corrections from subsequent classification refinement. Our solution employs a hybrid attention-recurrent architecture that synergizes sequential processing with graph structural priors.

\subsubsection{Task Formalization}
Given a set of \( n \) classified symbols \( \{s_i\}_{i=1}^n \), each with an associated sequence of traces, our goal is to predict directed edges \( e_{ij} \in \mathcal{R} \) representing spatial relationships between symbol pairs, where
\[
\mathcal{R} = \{\texttt{right}, \texttt{sup}, \texttt{sub}, \texttt{over}, \texttt{under}, \texttt{line\_start}\}.
\]
This is framed as a multi-label sequence-to-sequence problem, with each symbol \( s_i \) querying potential relations with every subsequent symbol \( s_j \) (\( j > i \)). 

To enforce a valid tree structure, each symbol \( s_j \) must receive exactly one incoming edge:
\[
\sum_{i=1}^{n} I(e_{ij} \neq \emptyset) = 1 \quad \forall j.
\]
The \texttt{line\_start} relation occurs exactly once per expression, ensuring a unique root, while each symbol can have multiple outgoing edges to subsequent symbols.

\subsubsection{Architectural Framework}  
The relation classification network predicts spatial relationships between symbols by processing sequential trace features through a streamlined architecture. The network first encodes an input sequence that integrates trace information along with classification features, augmented by a binary mask conveying world context, using a 2-layer bidirectional LSTM with a hidden size of 64. This produces contextualized representations of dimension 128 at each time step, followed by a dropout layer ($p=0.4$) to mitigate overfitting.

A subsequent multi-head attention mechanism (8 heads, embedding dimension 128) aggregates global contextual dependencies across the sequence. The resulting features are then passed to a two-stage fully-connected classifier. In the first stage, a linear projection maps the features to a 128-dimensional space, followed by layer normalization and a leaky ReLU activation. In the second stage, a linear transformation maps these features to the desired number of relation classes. At the prediction stage, a one-max operation is applied for specific relation categories (\texttt{right} and \texttt{line\_start}) to ensure that only one relation of that type is active per symbol. Finally, the network outputs log-probabilities over the relation classes using a log softmax activation.

Training is performed with epoch-level augmentation applied to both the input traces and the classification feature mask, initially incorporating synthetically generated data. Subsequent fine-tuning is carried out on real data without augmentation to further enhance accuracy. Ablation studies show that removing multi-head attention does only have a minor negative impact on relation classification accuracy, but it makes the training process more stable and converge faster.

\subsubsection{Results}

Our symbol relation module employs a hybrid attention-recurrent architecture to capture spatial dependencies between symbols, enhancing the structural interpretation of mathematical expressions. As evaluation metrics, we used relation classification accuracy (Rel), expression recognition accuracy (Exp), and structure recognition rate (Stru). Rel assesses the percentage of symbols with correctly classified relations, whereas Stru measures the number of expressions with both correct segmentation and relation detection evaluated on an expression level.

When trained solely on real data, the model achieves a Rel of 95.84\% (Exp: 53.05\%, Stru: 75.02\%). Incorporating the additional annotated data improves accuracy to 96.98\%, achieving a 1.14\% gain (Exp: 54.55\%, Stru: 77.94\%). Moreover, the integration of multi-head attention and batch normalization improved training stability, allowing us to achieve comparable results more quickly with a lower learning rate. These findings underscore the effectiveness of our approach in refining both relation prediction and overall expression accuracy. A revised relation prediction after the classification correction (see Sec. \ref{ssec:pipeline}) further improves expression accuracy by 0.14\%.

\subsection{Integrated Pipeline}  
\label{ssec:pipeline}  

Our system follows a multi-stage recognition paradigm enhanced with contextual refinement. The pipeline comprises:

\begin{enumerate}
    \item \textbf{Trace Segmentation}: Raw tracs are partitioned into symbol candidates through graph-based preprocessing and neural sequence modeling (Section~\ref{sec:segmentation}), achieving 99.7\% accuracy through spatial-temporal feature fusion.
    
    \item \textbf{Dual-Modal Classification}: Segmented symbols are recognized via parallel processing of trace trajectories (BiLSTM) and rendered glyphs (CNN) (Section~\ref{local_class}).
    
    \item \textbf{Primary Relation Prediction}: Spatial dependencies are established through attention-recurrent networks that jointly consider trace grouping geometry and preliminary class labels (Section~\ref{sec:relation}).
    
    \item \textbf{Global Context Correction}: A transformer module refines classifications using structural context from the partial SLG (Section~\ref{class_corr}).
    
    \item \textbf{Revised Relation Prediction}: Final dependencies are recomputed using corrected classifications, completing the SLG through constrained tree generation (Section~\ref{sec:relation}).
\end{enumerate}

\subsubsection{System-Wide Evaluation}
\label{ssec:evaluation }

Experiments on CROHME-2023 validate the system’s end-to-end efficacy using the evaluation metrics from \cite{xie2024local}: segmentation accuracy (Seg), symbol classification accuracy (Sym), relation accuracy (Rel), expression accuracy (Exp), and structural accuracy (Stru). Seg, Sym, and Rel assess trace grouping, symbol labeling, and relation detection at the symbol level, while Exp and Stru measure overall expression correctness and combined segmentation–relation performance at the expression level.  

Artificially annotating CROHME data via the proposed Automatic Annotation System enhanced every step of the recognition pipeline. Compared to the baseline model trained on real CROHME-2023 data, Seg increased from 98.22\% to 99.75\%, leading to higher Exp (53.05\% $\to$ 56.64\%) and Stru (75.02\% $\to$ 80.79\%).  

Classification accuracy also improved, with Sym increasing from 92.53\% (Exp: 53.05\%) to 95.01\% (Exp: 64.08\%). Adding a CNN pathway and correction module further raised Sym to 96.73\% (Exp: 74.27\%).  

Relation accuracy improved from 95.84\% (Exp: 53.05\%, Stru: 75.02\%) to 96.98\% (Exp: 54.55\%, Stru: 77.94\%) with the new approach. Combining these improvements produced a state-of-the-art model that outperforms graph-based baselines across all metrics (see Table~\ref{tab:task_accuracy}). Results were validated on the CROHME-2023 validation set for consistency with other methods.

\begin{table}[ht]
\centering
\caption{Task Performance on CROHME-2023 Validation Set. \\
Seg: Symbol Segmentation, \quad  Sym: Symbol Classification, \\
Rel: Relation Classification, \quad Exp: Expression Accuracy, \\
Stru: Structural Accuracy.}
\label{tab:task_accuracy}
\begin{tabular}{|l|c|c|c|c|c|}
\hline
\textbf{Method} & \textbf{Seg} & \textbf{Sym} & \textbf{Rel} & \textbf{Exp} & \textbf{Stru} \\ \hline

\textbf{Ours} (with gen. annotations) & \textbf{99.75\%} & \textbf{99.30\%} & \textbf{98.37\%} & \textbf{82.77\%} & \textbf{90.42\%} \\ \hline

GGM-EGAT (2024) \cite{xie2024local} & 98.39\% & 93.34\% & 93.69\% & 55.88\% & 80.73\%  \\ \hline

\textbf{Ours} (without gen. ann.) & 98.22\% & 92.53\% & 95.84\% & 
53.05\% & 75.02\% \\ \hline 

G2G (2021) \cite{chen2021edge} & 97.54\% & 90.88\% & 90.60\% & 45.30\% & 71.50\% \\ \hline
\end{tabular}
\end{table}

Table~\ref{tab:expression_accuracy} compares expression accuracy across CROHME variants. Our system achieves 74.14\% on CROHME-2023 without external resources, ranking competitively against methods using private data (e.g., Sunia) or LLMs (e.g., USTC-iFLYTEK). Crucially, our approach generates both LaTeX equations \textit{and} Stroke Label Graphs (SLG) that map recognized symbols to their input traces—providing structural annotations critical for interpretability and downstream applications.

\begin{table}[ht]
\centering
\caption{Expression accuracy comparison on different CROHME benchmarks. \\
\textit{Private}: models with no public implementation details;\\
\textit{SLG}: systems predicting a full Stroke Label Graph; \\
†: systems using additional private data; ‡: LLMs trained on external expressions.} 
\label{tab:expression_accuracy}
\begin{tabular}{|l|c | c|c c c|c|}
\hline
\textbf{System}                      & Private & SLG  & CROHME & CROHME & CROHME & Stru    \\
                                     &         &      & 2016   & 2019   & 2023   & 2019    \\ \hline
Sunia † \cite{crohme2023}            & x       &      & -      & 88.24\%& 82.34\%& 94.25\% \\ \hline
\textbf{Ours} (with gen. annotations)&         & x    & \textbf{80.52\%}& \textbf{83.63\%} & \textbf{74.14\%}   & \textbf{89.06\%} \\ \hline
YP\_OCR \cite{crohme2023}            & x       &      & -      & 84.74\%& 72.55\%& 92.66\% \\ \hline
USTC-iFLYTEK ‡ \cite{crohme2019}     & x       &      & -      & 80.73\%& -      & 91.49\% \\ \hline
Samsung R\&D † \cite{crohme2019}     & x       &      & 65.76\%& 79.82\%& -      & 89.32\% \\\hline
MyScript †‡ \cite{crohme2019}        & x       &      & 67.65\%& 79.15\%& -      & 90.66\% \\ \hline
PAL-v2 \cite{pal-v2}                 &         &      & 57.89\%& 62.89\%& -      & 79.15\% \\ \hline
GGM-EGAT \cite{xie2024local}         &         & x    & 56.67\%& 60.72\%& 55.30\%& 80.73\% \\ \hline
\textbf{Ours} (without gen. ann.)    &         & x    & 56.29\%& 54.63\%& 52.82\%& 77.15\% \\ \hline
G2G \cite{wu2021graph}               &         &      & 52.05\%& -      & 44.89\%& 71.50\% \\ \hline
\end{tabular}
\end{table}

\section{Discussion}
\label{sec:discussion}  

Our structural framework advances HMER through two key innovations: 1) automated structural annotation enabling data-efficient training of modular components, and 2) decoupled optimization of segmentation (99.75\% accuracy), classification (99.30\%), and relation prediction (98.37\%). These advancements are reflected in the high accuracy of symbol recognition and relation prediction, as demonstrated in Figure~\ref{fig:correct-predictions}, which shows correctly predicted examples:

\begin{figure}[H]
\vspace{-.5cm}
\centering
\caption{correctly predicted equations}
\vspace{-.3cm}
\begin{tabular}{cc}
\includegraphics[trim={0 7cm 0 8.5cm},clip,width=0.45\linewidth]{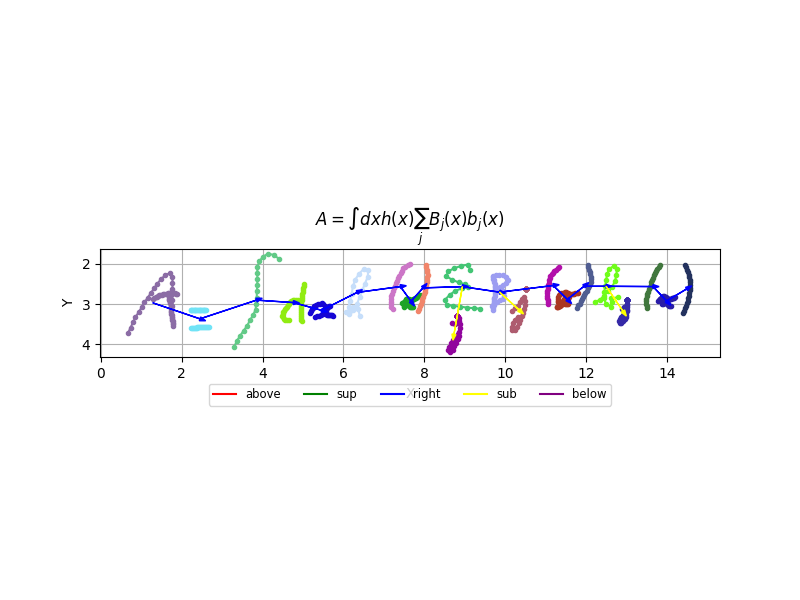} &
\includegraphics[trim={0 7cm 0 8.5cm},clip,width=0.45\linewidth]{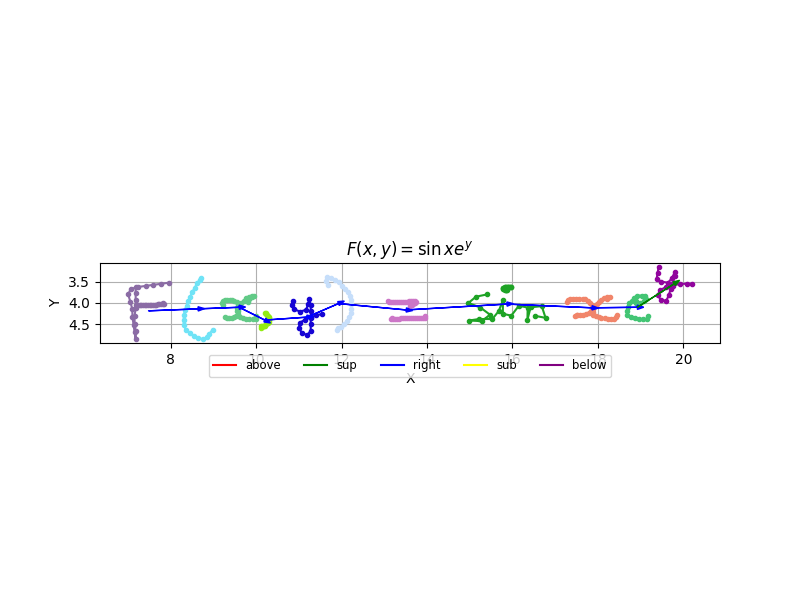} \\
(a) $A = \int d x h \left( x \right) \sum_j B_j \left( x \right) b_j \left( x \right)$ & (b) $F \left( x , y \right) = \sin x e^y$ \\
\end{tabular}
\label{fig:correct-predictions}
\vspace{-.5cm}
\end{figure}

\begin{figure}[H]
\vspace{-.5cm}
\centering
\caption{mispredicted equations}
\vspace{-.3cm}
\begin{tabular}{cc}
\includegraphics[trim={0 6.2cm 0 7.4cm},clip,width=0.45\linewidth]{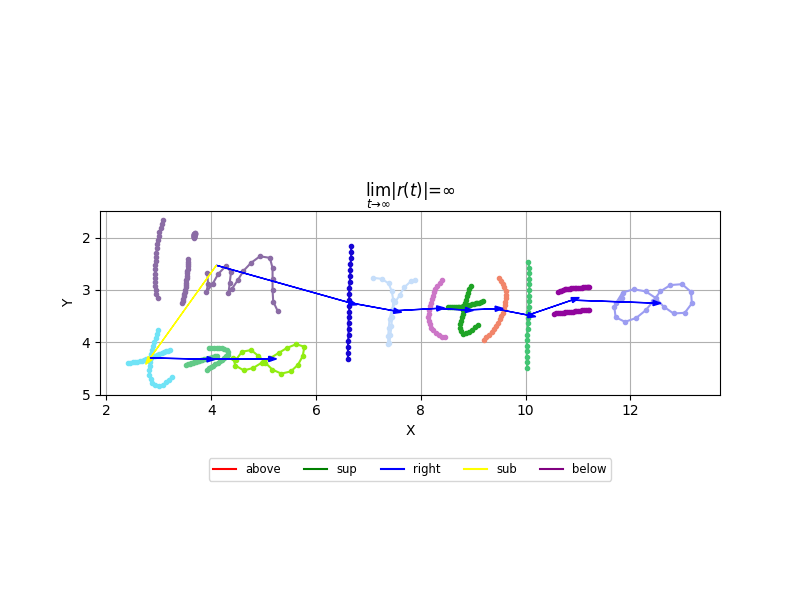} &
\includegraphics[trim={0 6.2cm 0 7.4cm},clip,width=0.45\linewidth]{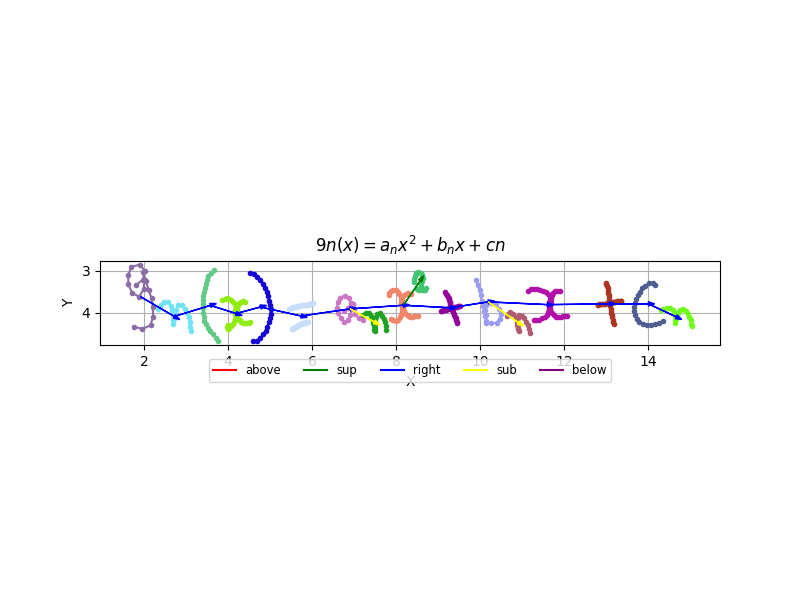} \\
(a) $\lim_{t \rightarrow \infty} | \textcolor{red}{r} \left( t \right) | = \infty$ &
(b) $\textcolor{red}{9n} \left( x \right) = a_n x^2 + b_n x + c n$ \\
correct: $\lim_{t \rightarrow \infty} | \gamma \left( t \right) | = \infty$ &
correct: $g_n \left( x \right) = a_n x^2 + b_n x + c_n$ \\
\end{tabular}
\label{fig:wrong-predictions}
\vspace{-.5cm}
\end{figure}

Our system achieves 74.14\% expression accuracy on CROHME-2023 while maintaining full structural interpretability through SLG generation. While this represents state-of-the-art performance for structurally grounded approaches, Figure~\ref{fig:wrong-predictions} highlights remaining challenges in distinguishing visually similar sub-expressions.

Despite the framework's strengths and advancements, we have identified several key areas where further improvements could enhance its broader applicability:  

\begin{itemize}  
    \item \textbf{Trace-Sharing Constraint:} The current 1:1 trace-to-symbol mapping poses challenges for recognizing cursive words or multi-letter symbols with shared traces. Addressing this would improve scalability to real-world equations with connected characters.  

    \item \textbf{Domain-Specific Scope:} While the framework performs well within CROHME's symbol set and clean, single-line equations, extending support for matrices, overlines, and additional symbols (e.g., \%, $\cdot$, $Y$) would make it more versatile for diverse real-world scenarios.  

    \item \textbf{Adaptation Efficiency:} Expanding the framework to new symbol sets or spatial relationships currently requires retraining with SLG annotations and potential pipeline adjustments. Streamlining this process would improve adaptability, similar to encoder-decoder systems that only need LaTeX-annotated data.  

    \item \textbf{Scalability Potential:} The sequential pipeline structure limits parallelization, creating challenges for handling large equations efficiently. Architectural adjustments could unlock greater scalability on modern hardware.  
\end{itemize}  

Promising directions to address these challenges include: 1) integrating word embedding techniques for improved symbol grouping to resolve trace-sharing issues, 2) expanding training to multi-domain datasets like MathWriting to broaden symbol coverage, and 3) developing synthetic augmentation strategies for multi-line and noisy expressions. Architectural refinements, such as increasing parallelism between pipeline stages, could enhance scalability while retaining interpretability. Additionally, semi-supervised learning approaches could reduce annotation dependence, further increasing the framework’s robustness and real-world applicability.

\newpage

\bibliographystyle{plain}
\bibliography{thesis}

\end{document}